\documentclass{ieeeaccess}
\usepackage{cite}
\usepackage{amsmath,amssymb,amsfonts}
\usepackage{algorithmic}
\usepackage{graphicx}
\usepackage{textcomp}

\makeatletter
\AtBeginDocument{\DeclareMathVersion{bold}
\SetSymbolFont{operators}{bold}{T1}{times}{b}{n}
\SetSymbolFont{NewLetters}{bold}{T1}{times}{b}{it}
\SetMathAlphabet{\mathrm}{bold}{T1}{times}{b}{n}
\SetMathAlphabet{\mathit}{bold}{T1}{times}{b}{it}
\SetMathAlphabet{\mathbf}{bold}{T1}{times}{b}{n}
\SetMathAlphabet{\mathtt}{bold}{OT1}{pcr}{b}{n}
\SetSymbolFont{symbols}{bold}{OMS}{cmsy}{b}{n}
\renewcommand\boldmath{\@nomath\mathversion{bold}}}
\makeatother

\def\BibTeX{{\rm B\kern-.05em{\sc i\kern-.025em b}\kern-.08em
    T\kern-.1667em\lower.7ex\hbox{E}\kern-.125emX}}

\usepackage{graphics} 

\usepackage[caption=false,font=normalsize,labelfont=sf,textfont=sf]{subfig}
\usepackage{stfloats}
\usepackage{float} 
\usepackage{here}

\usepackage[whole]{bxcjkjatype}

\usepackage{xspace}

\newcommand{\commentout}[1]{}

\newif\ifarxivnotice
\arxivnoticetrue
\ifarxivnotice
  \usepackage{eso-pic}
  \AddToShipoutPictureFG*{\AtPageUpperLeft{%
    \put(\LenToUnit{\dimexpr 1in+\hoffset+\oddsidemargin\relax},\LenToUnit{-8mm}){%
      \parbox[t]{130mm}{\raggedright\historyfont
        This work has been submitted to the IEEE for possible publication.\\
        Copyright may be transferred without notice, after which this version may no longer be accessible.}}}}
\fi

\begin{document}
\history{Date of publication xxxx 00, 0000, date of current version xxxx 00, 0000.}
\doi{10.1109/ACCESS.XXX}

\title{3D Pose Estimation in Projection Space via Pixel-level Visible Light Communication}
\author{\uppercase{Michikuni Eguchi}\authorrefmark{1,2},
\uppercase{Takefumi Hiraki}\authorrefmark{1,2}, \IEEEmembership{Member, IEEE}}

\address[1]{University of Tsukuba,
1-2 Kasuga, Tsukuba, Ibaraki 305-8550 Japan (e-mail: eguchi@mvml.slis.tsukuba.ac.jp, hiraki@slis.tsukuba.ac.jp)}
\address[2]{Cluster Metaverse Lab,
8-9-5 Nishigotanda, Shinagawa, Tokyo 140-0031 Japan}

\tfootnote{This work was partially supported by JST ASPIRE Grant Number JPMJAP2327.}

\markboth{Eguchi and Hiraki: 3D Pose Estimation in Projection Space via Pixel-level Visible Light Communication}{}

\corresp{Corresponding author: Takefumi Hiraki (e-mail: hiraki@slis.tsukuba.ac.jp).}

\begin{abstract}
Projector-based augmented reality enables natural interaction between users and devices by overlaying digital information onto the physical world through projected imagery.
A key technical requirement of such systems is accurate positional alignment between the projector and the devices.
Pixel-level Visible Light Communication (PVLC) achieves this alignment without external sensing infrastructure by embedding invisible spatial codes into each projected pixel through high-speed brightness modulation that preserves image quality.
However, the information that PVLC transmits is inherently planar.
PVLC has therefore been limited to devices moving on a flat surface and cannot estimate the 3D pose of devices in three-dimensional space, such as aerial robots.
To extend PVLC-based pose estimation to three dimensions, we formulate it as a Perspective-n-Point (PnP) problem.
The 2D pixel positions that four coplanar photosensors on the device receive via PVLC serve as the image points, and their known 3D layout provides the object points.
Experiments show that the proposed system estimates 3D pose with a latency of approximately 8~ms and a 3D position RMSE of 19.2~mm after Kalman filtering.
These results confirm the first realization of real-time 3D pose estimation coordinated with high-quality projected image content without external sensing infrastructure.
\end{abstract}

\begin{keywords}
Visible light communication, 3D pose estimation, perspective-n-point problem.
\end{keywords}

\titlepgskip=-21pt

\maketitle

\section{Introduction}
\label{sec:introduction}

\PARstart{I}{nteraction} systems based on augmented reality (AR) enable natural collaboration between users and devices by superimposing digital information onto the real world.
Among them, projector-based AR, also known as Spatial Augmented Reality (SAR)~\cite{bimber2005spatial}, overlays information directly onto real space with projected images.
Because users do not need to wear special devices such as head-mounted displays, SAR enables more natural interaction.
This technology has already produced many promising applications, including efficient teaching operations for industrial robots~\cite{10.1145/3371382.3378266, 10.1145/3351180.3351204}, tabletop interfaces using mobile robots~\cite{Rosenfeld2004-gd, 10.1145/3301275.3302326, 10.1145/3332165.3347911}, and interactive art~\cite{Darbar2019-kg}.

To seamlessly coordinate device behavior with projected image content, the system must obtain the position of the device relative to the projector with high accuracy.
Achieving such accuracy typically requires external sensing equipment such as cameras~\cite{9107397} or motion capture systems~\cite{furniss2000motion}, as well as geometric calibration between the projector and these sensors~\cite{lee2004automatic}.
This requirement increases setup complexity and limits practical deployment.

Pixel-level Visible Light Communication (PVLC)~\cite{kitamura2007KJ00007499118}, a projector-based visible light communication technique, has been proposed to address these challenges.
PVLC embeds invisible position codes into each pixel of the projected image through high-speed brightness modulation.
A photosensor on the device decodes the pixel coordinate at its location to obtain its own position within the image.
Because no external sensors or projector-sensor calibration is required, and because the embedded information is invisible to the human eye, PVLC enables device localization alongside high-quality projected content.
However, the projected image is inherently planar, and the position that a device obtains from it is limited to two dimensions.
PVLC has therefore been applied only to devices moving on a plane, such as ground robots~\cite{Hiraki2018-ri}, and cannot support devices that operate in three-dimensional space, such as aerial robots.

\Figure[t!](topskip=0pt, botskip=0pt, midskip=0pt)[width=0.99\textwidth]{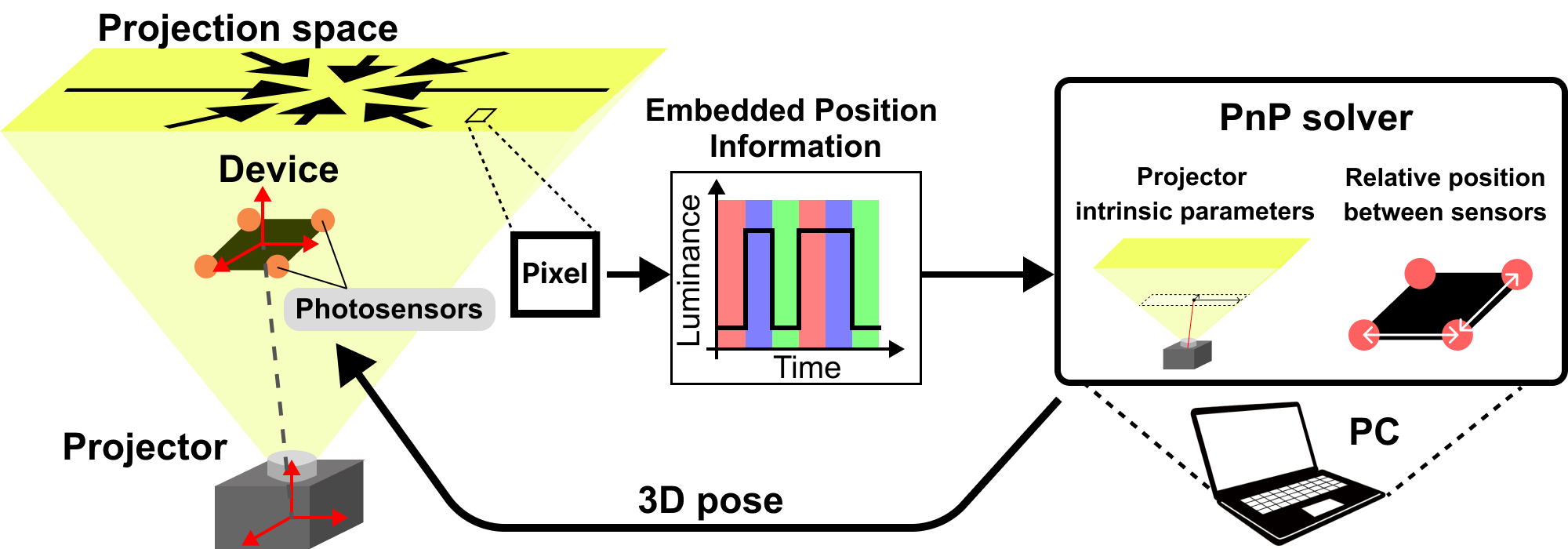}
{\textbf{Conceptual overview of the proposed 3D pose estimation system using Pixel-level Visible Light Communication (PVLC).} The projector embeds invisible position codes into each pixel. Multiple photosensors on the device decode their 2D pixel coordinates, which are combined with the projector's intrinsic parameters and the known 3D sensor geometry, and fed into a PnP solver to estimate the device's 3D pose.\label{fig:concept}}

To overcome this limitation, we extend PVLC-based localization from two dimensions to three by formulating it as a Perspective-n-Point (PnP) problem~\cite{Fischler1981-ce}.
Fig.~\ref{fig:concept} shows an overview of the proposed method.
The PnP problem is widely used for camera-based pose estimation.
In our formulation, the 2D pixel coordinates that multiple photosensors on the device decode via PVLC serve as the image points, and their known 3D layout provides the object points.
Solving this problem gives the 3D pose of the device with respect to the projector.
This approach enables 3D device control in coordination with high-quality projected images without projector-sensor calibration and opens up applications such as aerial robot navigation.

We conduct three experiments to validate the proposed method:
(1) evaluation of pose estimation accuracy under different movement speeds and photosensor layout sizes, with a camera-based AR marker method as a baseline,
(2) comparison of multiple PnP solvers to identify the most suitable one for the system, and
(3) investigation of the effectiveness of Kalman filter-based smoothing on estimation accuracy.
The results demonstrate that the system achieves centimeter-order position estimation accuracy with a latency of approximately 8~ms.

The main contributions of this paper are as follows:
\begin{itemize}
    \item The first application of PnP-based 3D pose estimation to PVLC. By solving the PnP problem with the 2D pixel coordinates decoded by multiple photosensors, the method extends PVLC-based localization from 2D to 3D.
    \item A content display region and a 3D pose tracking region that both cover the entire projection area at the same time. Because PVLC embeds the position codes into every pixel of the displayed content, the tracking region coincides with the content display region. Neither a dedicated tracking region nor an external sensor is required.
    \item Evaluation on a device moving at speeds from 50 to 400~mm/s. By combining the pose estimation with a Kalman filter, the method achieved a 3D position RMSE of 19.2~mm over all trajectories and movement speeds with the 80~mm photosensor layout.
\end{itemize}

\section{Related Work}
\label{sec:related_work}
Pose estimation using projected light determines the position and orientation of a target device with light sources fixed in the environment. This approach is widely used for high-precision indoor localization.
These methods can be broadly classified into two categories: methods that observe light projected onto a device with external cameras, and methods that employ visible light communication.

\subsection{Pose Estimation Method Using Cameras}
In external-camera-based methods~\cite{furniss2000motion,optitrack}, markers composed of retroreflective materials are attached to the device, and its pose is estimated by observing the markers with multiple cameras. This approach acquires the position of an object simply by attaching compact markers to it. It has therefore been widely adopted in human motion capture, which requires accurate measurement of numerous joint positions~\cite{davis1991gait,CHARLTON2004213}.
Structured-light methods project specific patterns, such as grids and dots, from a light source. They use the projected patterns as feature points and thus acquire positions without markers~\cite{Yuichi_TAMURA2012,6909541,6909459}.
Time-of-flight (ToF) cameras have also been used to estimate the pose of markerless targets from depth point clouds~\cite{park2025tof}.
These camera-based methods are well-established approaches for device pose estimation. However, to create applications that coordinate projected content with device movement, high-accuracy calibration between the observing camera and the projection display is required. In addition, the device must receive its position from the external camera system over a separate communication channel. This communication introduces non-negligible latency into the position update.

\subsection{Pose Estimation Method Using Visible Light Communication}
Visible light communication (VLC) has been actively studied for indoor positioning, and a wide range of algorithms have been proposed~\cite{Luo2017-vlc-survey}.
Projector-based VLC methods transmit position information through high-speed blinking of projector light, and photosensors on the device receive it for pose estimation. Various techniques have been proposed, including methods that project multiple Gray code patterns~\cite{lee2004automatic,10.1145/1095034.1095045,Le_Goc2016-dg,10.1145/1186562.1015738} and M-sequence patterns~\cite{10.1145/2501988.2502022}, as well as methods that alternately blink dozens of LEDs~\cite{1565405}. These methods enable high-speed position acquisition, but they degrade the native image quality of the projector by reducing resolution or sacrificing full-color display. This degradation diminishes the user experience in applications that coordinate projected content with device motion.

To address these limitations, we focus on PVLC~\cite{kitamura2007KJ00007499118}. PVLC uses the digital micromirror device (DMD) of a projector to rapidly modulate the brightness of the three primary color channels of every pixel in the projected image. This modulation transmits position information while preserving the perceived image quality. PVLC has been applied to various domains, including robot swarm control~\cite{Hiraki2018-ri,Hiraki2019-fn} and extension to infrared wavelengths for improved image contrast~\cite{Kamei2019-pilc}. However, all existing PVLC-based pose estimation methods remain limited to 2D localization, as the transmitted information inherently encodes only 2D pixel coordinates. While VLC-based 3D localization for unmanned aerial vehicle (UAV) swarms has been explored using ceiling-mounted LEDs~\cite{Xu2023-uav-vlc}, it does not address coordination with projector-displayed content. Spatial interactive applications such as aerial robot navigation require 3D pose estimation with PVLC, but no prior work has addressed this problem.

\subsection{Perspective-n-Point Formulation in Projector Systems}
In computer vision, the PnP problem~\cite{Fischler1981-ce} is a well-established framework for recovering 3D pose from 2D--3D point correspondences. Efficient solvers such as EPnP~\cite{Lepetit2009-yg} and the homography decomposition method~\cite{Faugeras1988MotionAS} are widely used in camera-based systems.

The same formulation appears in projector-based SAR. In this setting, a projector is treated as an inverse camera and registered onto a 3D physical object from 2D--3D correspondences between projector pixels and object points~\cite{raskar2002cartoon}. Furthermore, photosensors embedded in the target can acquire such correspondences automatically by receiving the projection patterns presented by the projector~\cite{lee2004automatic,10.1145/1186562.1015738}. Projector-based tracking extends this idea to moving display surfaces~\cite{10.1145/1095034.1095045}. It is the closest to our problem setting, as it localizes a device using a projector and photosensors mounted on the device. However, the limited refresh rate of conventional projectors requires such tracking to reserve dedicated regions for tracking patterns within the projected image, separate from the displayed content. These regions occupy part of the image without contributing to the content. Because tracking is lost once a photosensor leaves the tracking region, the movement speed of the device is also limited.

In this paper, we apply the PnP formulation to PVLC by treating photosensor positions as 3D reference points and the received pixel coordinates as their 2D projections. Because PVLC embeds the position codes at a modulation rate far above the flicker fusion threshold~\cite{Farrell1971-tz}, viewers do not perceive them. Every pixel carries both the content and the position codes. The entire projection area therefore serves as the trackable region, and there is no dedicated region for the device to leave. As a result, the system displays content and estimates the 3D pose of a device over the whole projection space at the same time, without additional sensing infrastructure.

\section{Method}
\label{sec:method}

\subsection{Overview}
Fig.~\ref{fig:concept} shows an overview of our proposed system. This system consists of a projector that transmits pixel-level spatial codes, and a photosensor device that receives the projected light to decode its 2D pixel coordinates. Multiple photosensors are mounted on the photosensor device, each of which obtains the 2D pixel coordinates of the projector pixel illuminating it. These 2D coordinates, together with the known 3D layout of the photosensors, are then input to a PnP solver to estimate the 3D pose of the device. The estimated pose is expressed as a translation vector relative to the projector coordinate system, whose origin is defined at the projector's optical center.

\subsection{Pixel-level Visible Light Communication}

\Figure[t](topskip=0pt, botskip=0pt, midskip=0pt)[width=0.99\columnwidth]{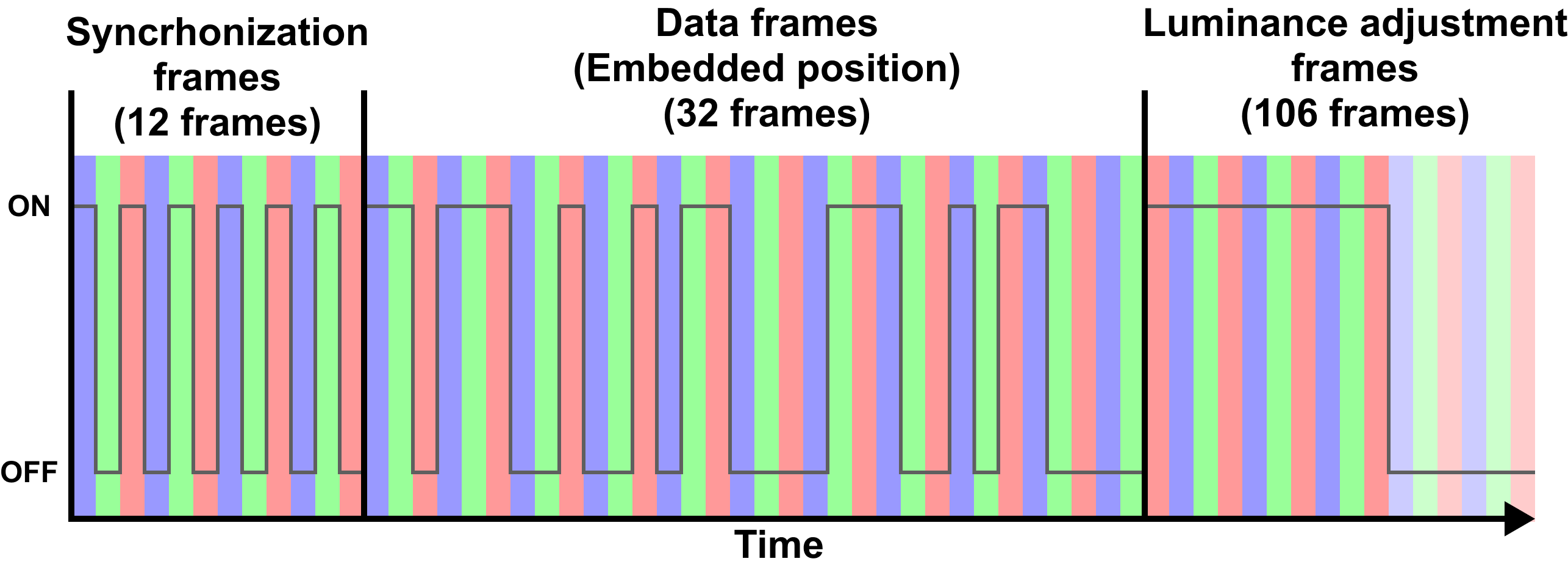}
{\textbf{Data structure of PVLC.}\label{fig:data_structure}}

\Figure[t](topskip=0pt, botskip=0pt, midskip=0pt)[width=0.99\columnwidth]{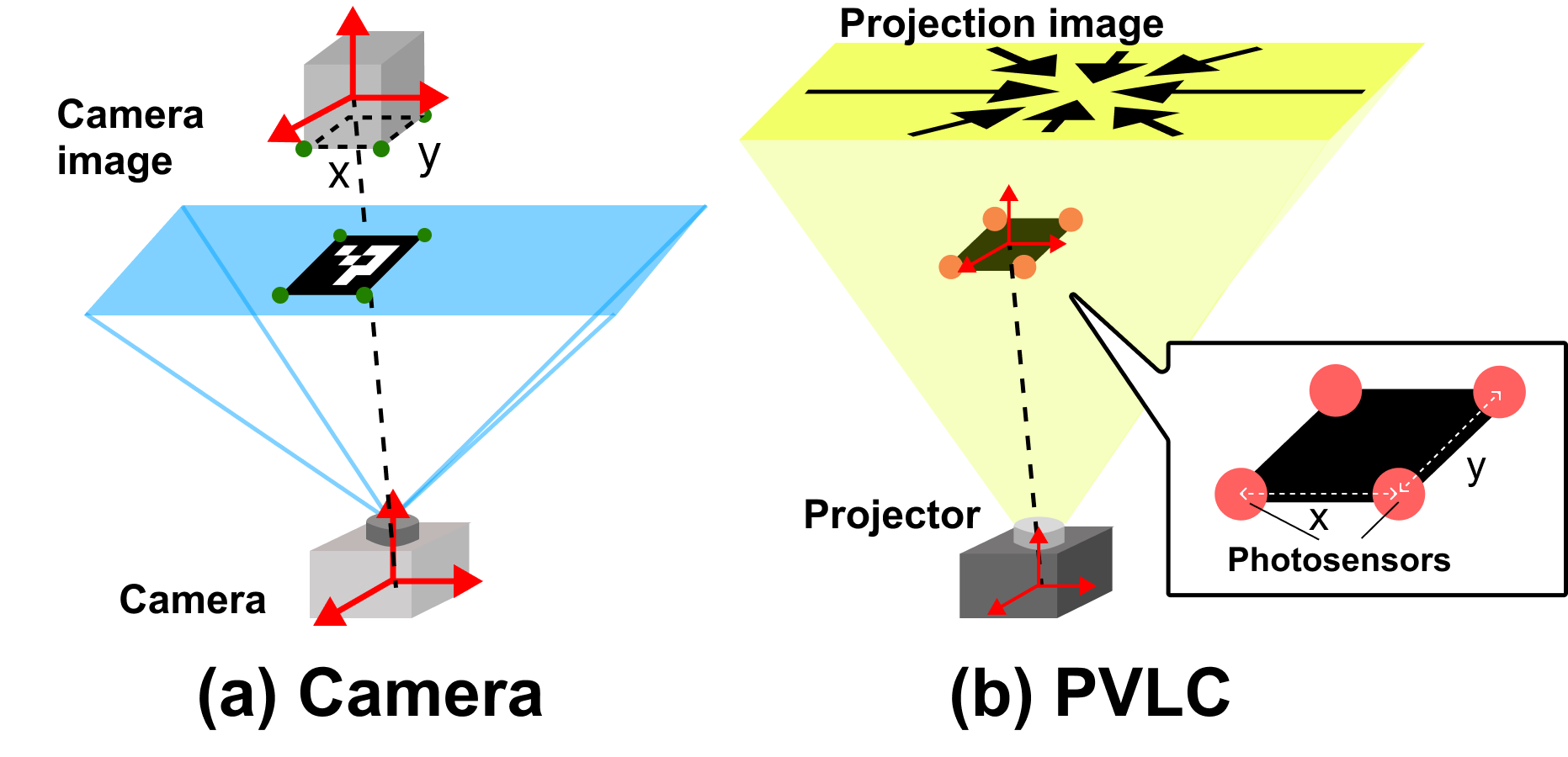}
{\textbf{Comparison of PnP-based 3D pose estimation in (a) a camera system and (b) the proposed PVLC system. In PVLC, (1) 3D marker points become photosensor positions, (2) the camera image plane becomes the projector pixel plane, and (3) 2D projected points become decoded pixel coordinates.} \label{fig:pnp}}

PVLC is a technology that transmits information by rapidly changing the brightness of each pixel in an image from a projector and receiving it with photosensitive elements such as photodiodes~\cite{Kimura2008-kd}. This method enables different data to be transmitted for each pixel. When the brightness changes at frequencies above 50--70 Hz, the human eye does not perceive the changes and sees an averaged static image~\cite{Farrell1971-tz}. PVLC utilizes this property to transmit information between devices while presenting images to humans.

The high-speed brightness changes in PVLC are achieved by the DMD built into Digital Light Processing (DLP) projectors. The DMD is an element in which microscopic mirrors are arranged two-dimensionally, with each mirror individually controllable. Each mirror switches the light on and off at high speed to control the brightness of its pixel.

Next, Fig.~\ref{fig:data_structure} shows the data structure of PVLC used in this research. This research utilizes the PVLC data structure proposed by Hiraki et al.\ for mobile robot control~\cite{Hiraki2018-ri}.
In this data structure, frames, which are the minimum units of transmitted information, are represented by the brightness of each RGB color space. These frames are composed of three types of frame groups: synchronization frames, data frames, and brightness adjustment frames.
The synchronization frame group uses 12 bits of frames to indicate the start of data transfer and embeds identical information across the entire screen to prevent reception errors.
The subsequent data frame group stores information such as position and control commands for the robot in a total of 32 bits.
The data frame group encodes data in Gray code to avoid reception errors due to light interference between multiple pixels. Gray code is an encoding scheme designed so that adjacent values differ by only one bit, minimizing the impact of mixing multiple pixels. In particular, this effect can effectively suppress the expansion of pose estimation error when the device is moving.
Finally, the brightness adjustment frame group consists of frames totaling 106 bits for generating images perceived by humans.
These three types of frame groups combine to form one transmission data unit, which is transmitted from the projector at a fixed cycle.
Each photosensor receives and decodes this transmission data unit following the procedure described in~\cite{Hiraki2018-ri}, thereby obtaining the 2D pixel coordinates of the projector pixel that is illuminating it. These pixel coordinates serve as the 2D input $\mathbf{p}_i$ to the PnP solver.

\Figure[t](topskip=0pt, botskip=0pt, midskip=0pt)[width=\textwidth]{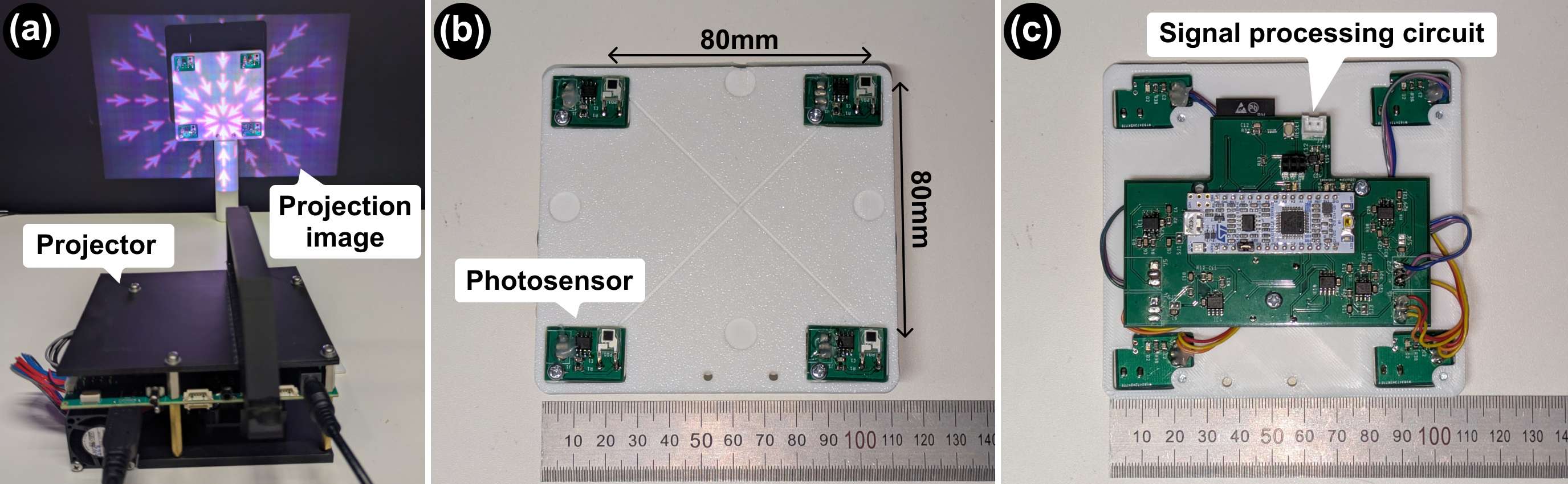}
{\textbf{Prototype sensor device.} (a) Overview of the sensor device receiving projected light from the projector. (b) Photosensor board with four photodiodes arranged in an 80 mm square on the same plane. (c) Signal processing circuit consisting of a current-to-voltage conversion circuit, amplification circuit, comparator circuit, and microcontroller.\label{fig:Implementation}}

\subsection{3D Pose Estimation Using PnP Problem}
The PnP problem is the problem of estimating the pose of a camera from the 3D coordinates $\mathbf{P}_i\in\mathbb R^{3}$ of multiple points $i$ in a specific marker coordinate system and the 2D pixel positions $\mathbf{p}_i\in\mathbb R^{2}$ corresponding to those points, and is widely used in the computer vision field.
The PnP problem is based on the following equation:

\begin{equation}
    s \: \mathbf{p}_i = \mathbf{A} \begin{bmatrix} \mathbf{R} | \mathbf{t} \end{bmatrix} \mathbf{P}_i
\end{equation}
where $s$ is a scale factor, $\mathbf{A}$ is the camera's intrinsic parameter measured in advance, and $\mathbf{R}\in\mathbb R^{3\times3}$ and $\mathbf{t}\in\mathbb R^{3\times1}$ are the rotation matrix and translation vector of the camera with respect to the marker coordinate system, respectively. In this equation, $\mathbf{p}_i$ and $\mathbf{P}_i$ are expressed in homogeneous coordinates. In the PnP problem, this equation is constructed for all point groups $i$, and the rotation matrix $\mathbf{R}$ and translation vector $\mathbf{t}$ are calculated by integrating these equations.
In the PVLC system, $\mathbf{A}$ corresponds to the intrinsic parameter matrix of the projector, including focal length and principal point. These parameters were obtained using the projector--camera calibration system proposed in~\cite{9107397}.

Normally, the PnP problem requires at least six 2D-3D correspondence point groups to calculate a unique rotation matrix and translation vector. On the other hand, this number of point groups can be reduced by establishing new constraints. For example, the Homography Decomposition method~\cite{Faugeras1988MotionAS}, which assumes that all point groups are placed on a plane, can obtain relatively accurate solutions with a minimum of four points.
Therefore, it is actively used as a pose estimation method for AR markers with four corners.
The correct solution is identified using the constraint that the sensor device is always oriented toward the projector.

The proposed method solves the PnP problem using the 2D pixel positions that multiple photosensors obtain from the projected image. It then calculates the 3D relative coordinates between the photosensor group and the projector.
Fig.~\ref{fig:pnp} shows a schematic diagram of 3D position estimation using the PnP problem solution in camera and PVLC systems.
The three main differences from camera usage examples are as follows:
First, the relative coordinates of point groups in 3D space are treated as relative coordinates between photosensors in PVLC.
Second, the camera image plane is replaced by the pixel plane of the projector.
Third, the pixel positions of points on the 2D image are based on the pixel positions of the projector's projected image captured by the photosensors in PVLC.
In this method, pose estimation of the device is performed by placing at least four photosensors on a plane, based on the aforementioned characteristics of the PnP problem. The 3D coordinates $\mathbf{P}_i$ of each photosensor are pre-defined based on the known geometric layout of the sensor board, in which four photosensors are arranged in an 80 mm square.

\subsection{Implementation}\label{sec:implementation}

We constructed a system as shown in Fig.~\ref{fig:Implementation} as a prototype for verifying the proposed position estimation method.
In this system, we used a DLP projector (DLP LightCrafter 4500, Texas Instruments) for PVLC operation.
From this projector, invisible data with the structure shown in Fig.~\ref{fig:data_structure} is transmitted at a resolution of 1140 $\times$ 912 per pixel at a 125 Hz cycle.

Next, a photosensor circuit and its signal processing circuit were installed on the sensor device.
A total of four photosensors were placed on the same plane in an 80 mm square arrangement for the operation of the proposed method (Fig.~\ref{fig:Implementation}(b)).
The photosensor circuit consists of a photodiode (S2506-02, Hamamatsu Photonics) and an operational amplifier (OPA2353UA, Texas Instruments), which converts the current output from the photodiode to voltage through a current-to-voltage conversion circuit and transmits it to the signal processing circuit.
The converted voltage is then amplified through an amplification circuit and converted to a digital signal through a comparator circuit. Variable resistors are attached to the amplification circuit and comparator circuit, and the received light can be correctly processed by adjusting the resistance values according to the operating environment.
In addition, a high-pass filter is incorporated into the signal processing circuit to suppress low-frequency ambient light from indoor lighting such as fluorescent lamps.
The digital signal of the projected light is transmitted to a microcontroller (NUCLEO-F303K8, STMicroelectronics), which performs decoding of the received data.
The position data obtained by the microcontroller is continuously transmitted to a PC (ThinkPad X13 Gen 2, Lenovo, CPU: AMD Ryzen 5 PRO 4650U, RAM: 16GB) via serial communication, and the 3D position of the device is estimated by solving the PnP problem.
The PnP problem solution was implemented using the Python API of the Homography Decomposition method~\cite{Faugeras1988MotionAS} provided by OpenCV~\cite{opencv_library}.

\section{Experiments}
\label{sec:experiments}
In this paper, we verified the 3D position estimation accuracy and its computation time to confirm the effectiveness of 3D position estimation in PVLC using the PnP-based pose estimation approach.

\subsection{Setup}
\subsubsection{Experiment System}
To evaluate the position estimation system, we built the experimental setup shown in Fig.~\ref{fig:experiment_system}.
We mounted the sensor device on the end effector of a manipulator (xArm5, UFACTORY), which moved it to and held it at specified positions. The manipulator has a positioning repeatability of $\pm$0.1 mm. Given this precision, we used the device position computed from the manipulator's kinematics as the ground truth. An AR marker placed at the manipulator base served as the origin of the position estimation system. A camera (Intel RealSense D415, Intel) attached to the projector side detected this marker. We used the detected marker pose to transform the projector coordinate system into the arm coordinate system. As shown in Fig.~\ref{fig:experiment_system}, all experiments were conducted in an ordinary indoor room with the room lights on.

\subsubsection{Evaluation Metrics}
In this experiment, we used Root Mean Square Error (RMSE) and Standard Deviation (SD) as metrics for evaluating the accuracy of estimated poses. For position, the Euclidean distance between the true position and the estimated position was used as the error. To confirm the change in estimation error with respect to the projector light irradiation direction, errors in 3D ($x,y,z$ space) and 2D ($x,y$ plane) were evaluated separately. For orientation, after calculating the difference between the true angle and the estimated angle in each of the roll, pitch, and yaw directions, the Euclidean norm value combining these three components was used as the error.

\Figure[t!](topskip=0pt, botskip=0pt, midskip=0pt)[width=0.99\columnwidth]{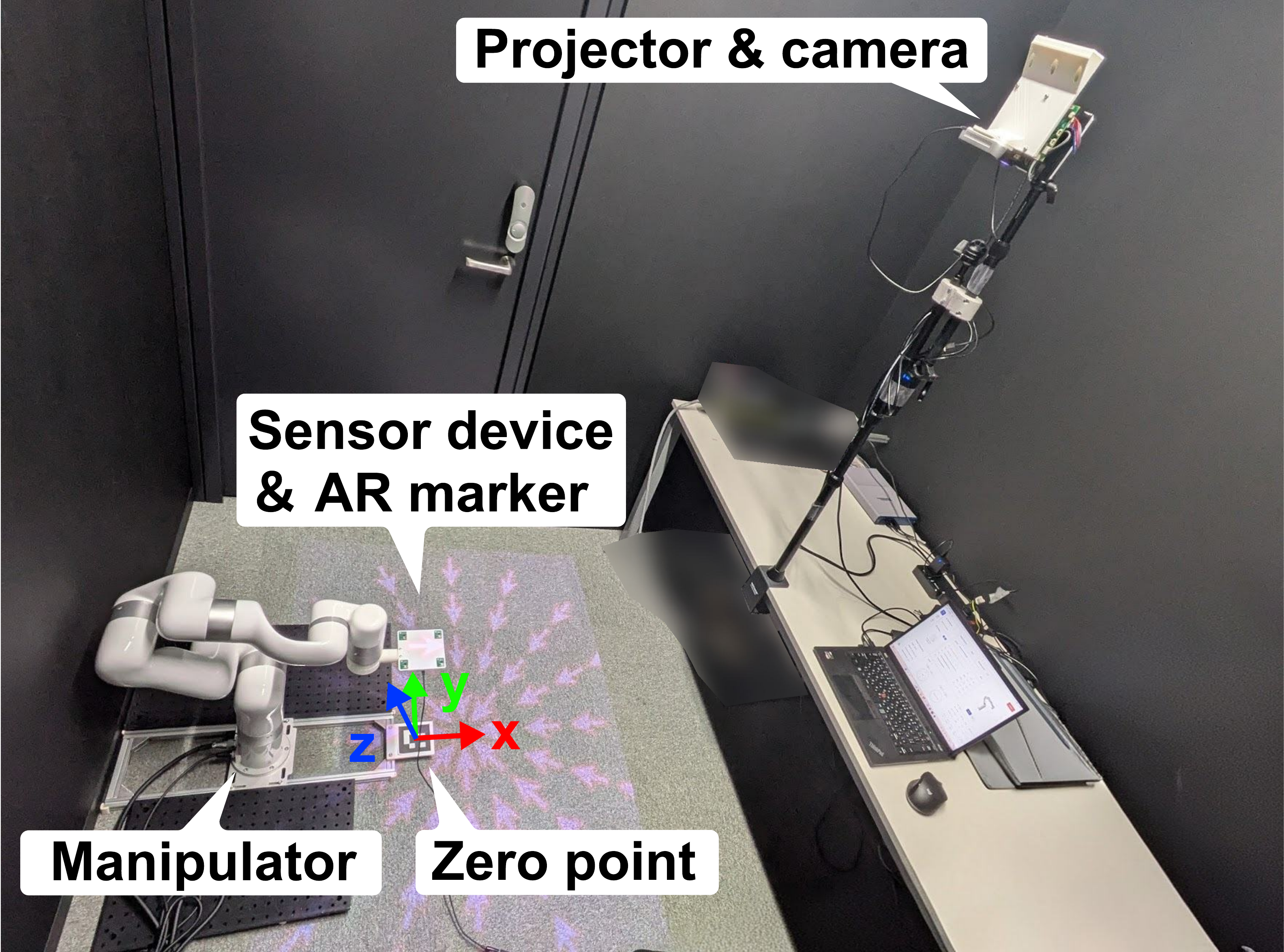}
{\textbf{Experimental system.} The sensor device is mounted on a robotic manipulator (xArm5) that provides ground-truth positions. An AR marker at the manipulator base defines the coordinate origin.\label{fig:experiment_system}}

\subsubsection{Comparison Method}
As a comparison method, we used camera-based AR marker pose estimation. We attached an AR marker (ArUco~\cite{GarridoJurado2014aruco}) to the sensor device and captured it with the projector-side camera described above. The ArUco module of OpenCV estimated the marker pose from the captured images. We used a 4 $\times$ 4 ArUco marker (\texttt{DICT\_4X4\_50} in OpenCV). The camera captured color images at 1920 $\times$ 1080 pixels and 30 fps with auto exposure.

\subsubsection{Experimental Conditions}\label{sec:conditions}
We evaluated the pose estimation accuracy of both methods while varying the layout size and the movement speed. The photosensors were arranged in squares with side lengths of 80 mm and 160 mm, and the AR markers had the same side lengths. Both methods therefore solve a PnP problem with four coplanar points at the same spacing. The manipulator moved the sensor device along the four trajectory patterns shown in Fig.~\ref{fig:error_dynamic} at 50, 100, 200, and 400 mm/s, and five laps were performed under each condition. To compare the two methods fairly despite their different output rates, we downsampled all estimates to a 40 ms interval. For each condition, we calculated the RMSE and SD for each trajectory pattern and averaged them over the four patterns.

\subsection{Comparison with AR Markers}\label{sec:localize_dynamic}
Table~\ref{tab:error_dynamic} shows the pose estimation error for each condition in Section~\ref{sec:conditions}, and Figs.~\ref{fig:error_speed} and~\ref{fig:error_speed_angle} show the 3D position error and the angle error versus movement speed, respectively.
From 50 mm/s to 400 mm/s, the 3D position error of the proposed method increased by approximately 33\% with the 160 mm layout and 20\% with the 80 mm layout. The angle error remained almost constant regardless of speed. The AR marker showed smaller errors than the proposed method at low speeds, but its 3D position error increased by a factor of three to four over the same range. Its angle error increased in a similar manner. As the speed increased, the advantage of the AR marker diminished. At the highest speed, the proposed method with the 160 mm layout achieved a lower 3D position error than both AR markers.

Enlarging the photosensor layout from 80 mm to 160 mm also substantially improved the accuracy of the proposed method. At 50 mm/s, the 3D position error decreased by approximately 40\% and the angle error by approximately 65\%. Enlarging the AR marker reduced its 3D position error by approximately 30\%.
We discuss the effect of the layout size in Section~\ref{subsec:theoretical_bounds} and the comparison with the AR marker, including the speed dependence, in Section~\ref{subsec:speed_dependence}.

We also compared the pose estimation latency of the two methods. The average latency was 40.2 ms for the AR marker and 8.0 ms for the proposed method.



\begin{table*}
\caption{\textbf{Pose estimation error values of the proposed PVLC and the AR marker for different movement speeds and marker sizes. For each condition, the values are averaged over all trajectory patterns. Values are presented as RMSE $\pm$ SD.}}
\label{tab:error_dynamic}
\centering
\setlength{\tabcolsep}{4pt}
\begin{tabular}{c|cc|cc|cc|cc|}
     & \multicolumn{2}{c|}{\textbf{50 mm/s}} & \multicolumn{2}{c|}{\textbf{100 mm/s}} & \multicolumn{2}{c|}{\textbf{200 mm/s}} & \multicolumn{2}{c|}{\textbf{400 mm/s}} \\
     & \textbf{3D [mm]} & \textbf{Angle [deg]} & \textbf{3D [mm]} & \textbf{Angle [deg]} & \textbf{3D [mm]} & \textbf{Angle [deg]} & \textbf{3D [mm]} & \textbf{Angle [deg]} \\ \hline
    \textbf{PVLC (Ours) 160 mm} & $13.5 \pm 7.3$ & $7.4 \pm 5.5$ & $14.2 \pm 7.9$ & $7.4 \pm 5.5$ & $15.3 \pm 8.5$ & $7.2 \pm 5.3$ & $17.9 \pm 10.0$ & $6.8 \pm 4.9$ \\
    \textbf{PVLC (Ours) 80 mm} & $23.4 \pm 13.9$ & $21.6 \pm 15.5$ & $25.1 \pm 15.2$ & $21.3 \pm 15.4$ & $25.3 \pm 14.9$ & $21.9 \pm 15.8$ & $28.1 \pm 17.1$ & $22.9 \pm 16.2$ \\
    \textbf{AR 160 mm} & $7.0 \pm 2.6$ & $2.4 \pm 0.7$ & $8.0 \pm 2.8$ & $2.5 \pm 0.9$ & $14.0 \pm 5.3$ & $3.4 \pm 1.5$ & $26.7 \pm 13.5$ & $6.1 \pm 2.9$ \\
    \textbf{AR 80 mm} & $9.8 \pm 3.7$ & $4.5 \pm 3.3$ & $11.5 \pm 4.4$ & $4.7 \pm 3.3$ & $19.7 \pm 9.0$ & $7.0 \pm 4.2$ & $30.3 \pm 17.6$ & $11.7 \pm 7.0$ \\
\end{tabular}
\end{table*}

\Figure[t!](topskip=0pt, botskip=0pt, midskip=0pt)[width=\textwidth]{figures/PVLC_dynamic.pdf}
{\textbf{Four trajectory patterns used in the dynamic experiments and examples of the pose estimation results of the proposed PVLC and the AR marker. All examples were obtained with a marker size of 80 mm at a movement speed of 200 mm/s.}\label{fig:error_dynamic}}

\Figure[t!](topskip=0pt, botskip=0pt, midskip=0pt)[width=0.99\columnwidth]{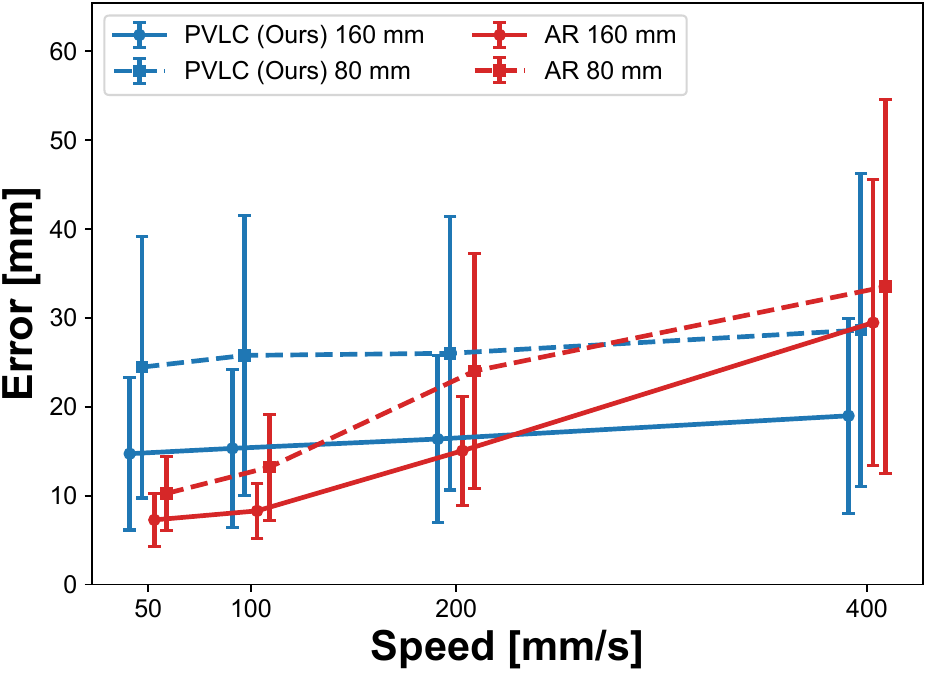}
{\textbf{3D position error (RMSE $\pm$ SD) versus movement speed for the proposed PVLC and the AR marker with marker sizes of 160 mm and 80 mm.}\label{fig:error_speed}}

\Figure[t!](topskip=0pt, botskip=0pt, midskip=0pt)[width=0.99\columnwidth]{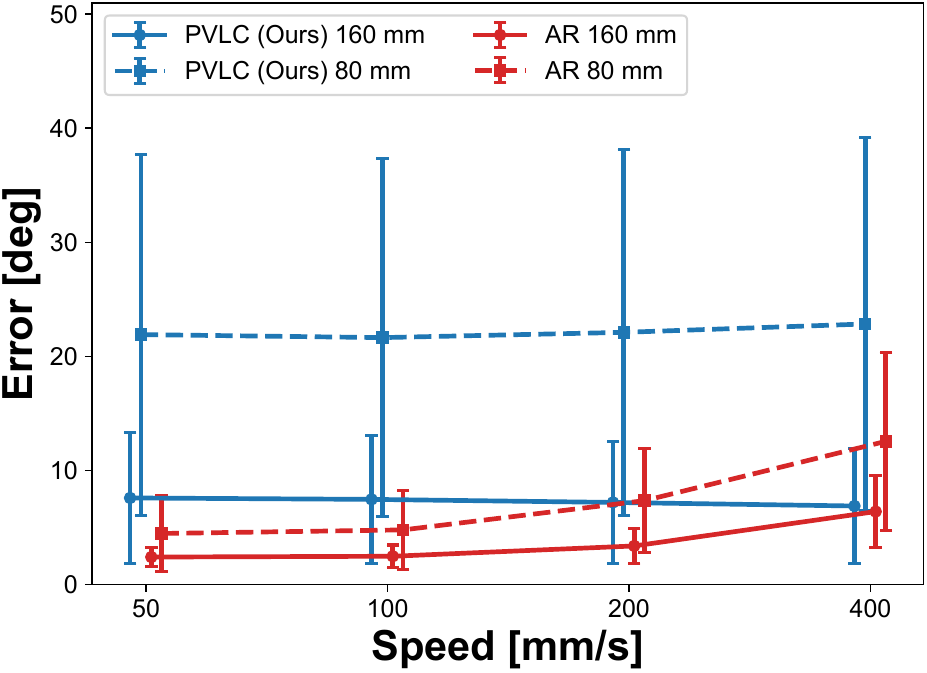}
{\textbf{Angle error (RMSE $\pm$ SD) versus movement speed for the proposed PVLC and the AR marker with marker sizes of 160 mm and 80 mm.}\label{fig:error_speed_angle}}

\subsection{Comparison of PnP Solvers}\label{subsec:exp_pnp}
In the proposed system, the number of photosensors is subject to constraints in terms of cost and device size, making it impractical to arbitrarily increase the number of correspondence points. Therefore, selecting a PnP solver that achieves stable accuracy with a limited number of points is critical. To this end, we compared the pose estimation accuracy and processing time of four PnP solvers to identify the most suitable solver for our system. The solvers used were the following four, including the Homography Decomposition method used in the experiment in Section~\ref{sec:localize_dynamic}:
\begin{itemize}
    \item \textbf{Homography Decomposition (HD)}~\cite{Faugeras1988MotionAS}:
        A method that estimates the homography matrix from four or more planar correspondence points and decomposes it to determine the object pose. It is a classical approach for planar objects and is widely used for AR marker pose estimation.

    \item \textbf{IPPE}~\cite{collins2014infinitesimal}:
        A method that determines the object pose by assuming infinitesimal rotation of a planar object with four or more correspondence points. It has been reported to be superior to the Homography Decomposition method in both estimation accuracy and processing speed.

    \item \textbf{P3P}~\cite{1217599}:
        A method that determines pose from three correspondence points and generally returns up to four candidate solutions. The most plausible solution is selected based on the smallest reprojection error. In this experiment, all combinations of three photosensors from the four available were evaluated, and the one with the smallest reprojection error was selected.

    \item \textbf{EPnP}~\cite{Lepetit2009-yg}:
        An algorithm that efficiently solves the PnP problem for a large number of correspondence points. It is widely adopted as a PnP solver in real-time applications.
\end{itemize}
For evaluation, the sensor device with the 80 mm photosensor layout was moved along the trajectory shown in Fig.~\ref{fig:error_pnp} for five laps at each speed (50, 100, 200, and 400 mm/s). The acquired photosensor data were processed by each solver, and the errors were aggregated over all movement speeds. Fig.~\ref{fig:error_pnp} shows an example of the estimated trajectories at a movement speed of 100 mm/s.

Table~\ref{tab:error_pnp} shows the pose estimation error and average solver processing time for each PnP solver. First, from the perspective of error magnitude, the most accurate among the compared solvers were the Homography Decomposition method and IPPE, which target planes, both showing similar error magnitudes. Next was P3P, where in some cases the estimated values deviated significantly from the reference trajectory. The least accurate was EPnP, where although the solver occasionally produced values approaching the reference trajectory, most results deviated significantly from it. Next, from the perspective of processing time, all solvers were fast at less than 1 ms, having almost no impact on the system's position update latency of approximately 8 ms.

\begin{table*}
\caption{\textbf{Pose estimation error values and solver processing time comparing PnP solvers. The errors were aggregated over five laps at all movement speeds along the trajectory shown in Fig.~\ref{fig:error_pnp} with the 80 mm photosensor layout. Values are presented as RMSE $\pm$ SD.}}
\label{tab:error_pnp}
\centering
\begin{tabular}{c|c|c|c|}
     & \textbf{3D error [mm]} & \textbf{Angle error [deg]} & \textbf{Processing time [ms]}
     \\ \hline
    \textbf{HD} & $23.0 \pm 13.2$ & $\mathbf{22.2 \pm 15.8}$ & $0.22 \pm 0.06$
    \\
    \textbf{IPPE} & $\mathbf{22.8 \pm 13.2}$ & $24.2 \pm 16.4$ & $\mathbf{0.01 \pm 0.01}$
    \\
    \textbf{P3P} & $29.6 \pm 18.2$ & $24.4 \pm 15.1$ & $0.02 \pm 0.00$
    \\
    \textbf{EPnP} & $121.0 \pm 62.3$ & $50.7 \pm 15.4$ & $0.03 \pm 0.01$
    \\
\end{tabular}
\end{table*}

\Figure[t!](topskip=0pt, botskip=0pt, midskip=0pt)[width=0.95\textwidth]{figures/PVLC_PnP.pdf}
{\textbf{The movement trajectory and the pose estimation results comparing PnP solvers. The example was obtained with the 80 mm photosensor layout at a movement speed of 100 mm/s.}\label{fig:error_pnp}}

\subsection{Result of Smoothing Process with Kalman Filter}
\label{sec:exp_kalman}
We investigated whether the pose estimation results of the proposed system could be improved by filter processing, and performed smoothing processing on the pose estimation results using the Kalman Filter (KF)~\cite{10.1115/1.3662552}.
We used the constant acceleration model~\cite{jondhale2018tracking} as the sensor's motion model and constructed the Kalman filter based on the following mathematical model.
In this model, the state $\boldsymbol{x}_k$ at time $k$ consists of the position $\boldsymbol{p}_k$, the rotation vector $\boldsymbol{r}_k$, and their first and second time derivatives:
\begin{equation}
    \boldsymbol{x}_k =
    \begin{bmatrix}
    \boldsymbol{p}_k^\top ~
    \boldsymbol{r}_k^\top ~
    \dot{\boldsymbol{p}}_k^\top ~
    \dot{\boldsymbol{r}}_k^\top ~
    \ddot{\boldsymbol{p}}_k^\top ~
    \ddot{\boldsymbol{r}}_k^\top
    \end{bmatrix}^\top \in \mathbb R^{18}.
\end{equation}
Here, $\boldsymbol{p}_k\in\mathbb R^3$ is the 3D position, and $\boldsymbol{r}_k\in\mathbb R^3$ is the deviation of the estimated orientation from the reference orientation, expressed as a rotation vector.
The state equation and observation equation are expressed as follows, and are updated with Gaussian noise $\boldsymbol{\omega}$ and $\boldsymbol{\nu}$ represented by covariance matrices $Q_{\omega}\in \mathbb R^{18\times18}$ and $R_{\nu}\in \mathbb R^{6\times6}$:
\begin{equation}
    \boldsymbol{x}_{k+1}
        = \begin{bmatrix}
        \mathbf{I}_6 & \Delta t\,\mathbf{I}_6 & \tfrac12 \Delta t^2\,\mathbf{I}_6 \\
        \mathbf{0}   & \mathbf{I}_6           & \Delta t\,\mathbf{I}_6           \\
        \mathbf{0}   & \mathbf{0}             & \mathbf{I}_6
    \end{bmatrix}\,\boldsymbol{x}_k + \boldsymbol{\omega}_k,
        \quad
        \boldsymbol{\omega}_k \sim \mathcal{N}\bigl(\boldsymbol{0}, Q_{\omega}\bigr).
\end{equation}
\vspace{-2mm}
\begin{equation}
    \boldsymbol{z}_k
    = \begin{bmatrix}
    \mathbf{I}_6 & \mathbf{0} & \mathbf{0}
    \end{bmatrix}\,\boldsymbol{x}_k + \boldsymbol{\nu}_k,
    \quad
    \boldsymbol{\nu}_k \sim \mathcal{N}\bigl(\boldsymbol{0}, R_{\nu}\bigr),
\end{equation}
where $\Delta t$ is the update interval of the filter, set to 40 ms to match the evaluation interval, $\mathbf{I}_6$ is the $6\times6$ identity matrix, and $\mathbf{0}$ is the $6\times6$ zero matrix. We set the process noise covariance to
\begin{equation}
    Q_{\omega} = \mathrm{diag}\Bigl(\tfrac{\Delta t^3}{6}\mathbf{1}_6^\top,~ \tfrac{\Delta t^2}{2}\mathbf{1}_6^\top,~ \Delta t\,\mathbf{1}_6^\top\Bigr)^2,
\end{equation}
where $\mathbf{1}_6$ is the six-dimensional vector of ones. The observation noise standard deviations were 0.02 m along the $x$ and $y$ axes, 0.16 m along the $z$ axis, and 0.8 rad for each rotation vector component, giving $R_{\nu} = \mathrm{diag}(0.02^2, 0.02^2, 0.16^2, 0.8^2, 0.8^2, 0.8^2)$. We applied the filter offline to the recorded pose estimates and determined the observation noise standard deviations by grid search. The selected value along $z$ was larger than those along $x$ and $y$ because the PnP solution has a larger error in depth than in the lateral directions.

For evaluation, we used the data of all trajectory patterns and movement speeds with the 80 mm photosensor layout in Section~\ref{sec:localize_dynamic}. Table~\ref{tab:error_kalman} and Fig.~\ref{fig:error_kalman} show the pose estimation error and trajectory before and after processing with the Kalman filter, respectively. As a result of the smoothing process, we confirmed that the 3D position error could be improved from 25.4 mm in the raw signal to 19.2 mm. The 2D position error and angle error also decreased from 10.2 mm to 9.0 mm and from 22.0$^{\circ}$ to 16.0$^{\circ}$, respectively. Moreover, it was confirmed from the estimated trajectory (Fig.~\ref{fig:error_kalman}) that the vibration in the z-axis direction could be greatly smoothed. From these findings, smoothing processing including the Kalman filter can be said to work effectively on the proposed system.

To check whether the filter correctly estimates its own uncertainty, Fig.~\ref{fig:kalman_3sigma} shows the estimation error after filtering for one lap. The shaded region indicates the $\pm3\sigma$ bounds, where $\sigma$ is the standard deviation of the estimation error predicted by the filter. We obtained $\sigma$ from the diagonal elements of the error covariance $P_k$. If the uncertainty estimate is correct, the actual error should stay within these bounds most of the time. For all six components, the error remained within the bounds throughout the lap.

\begin{table*}
\caption{\textbf{Pose estimation error values of the raw PVLC signal and the Kalman-filtered signal with the 80 mm photosensor layout, aggregated over all trajectory patterns and movement speeds. Values are presented as RMSE $\pm$ SD.}}
\label{tab:error_kalman}
\centering
\begin{tabular}{c|c|c|c|}
     & \textbf{3D error [mm]} & \textbf{2D ($x,y$) error [mm]} & \textbf{Angle error [deg]}
     \\ \hline
    \textbf{Raw signal} & $25.4 \pm 15.4$ & $10.2 \pm 6.6$ & $22.0 \pm 15.9$
    \\
    \textbf{Kalman-filtered signal} & $\mathbf{19.2 \pm 12.1}$ & $\mathbf{9.0 \pm 5.9}$ & $\mathbf{16.0 \pm 9.2}$
    \\
\end{tabular}
\end{table*}

\Figure[t!](topskip=0pt, botskip=0pt, midskip=0pt)[width=\textwidth]{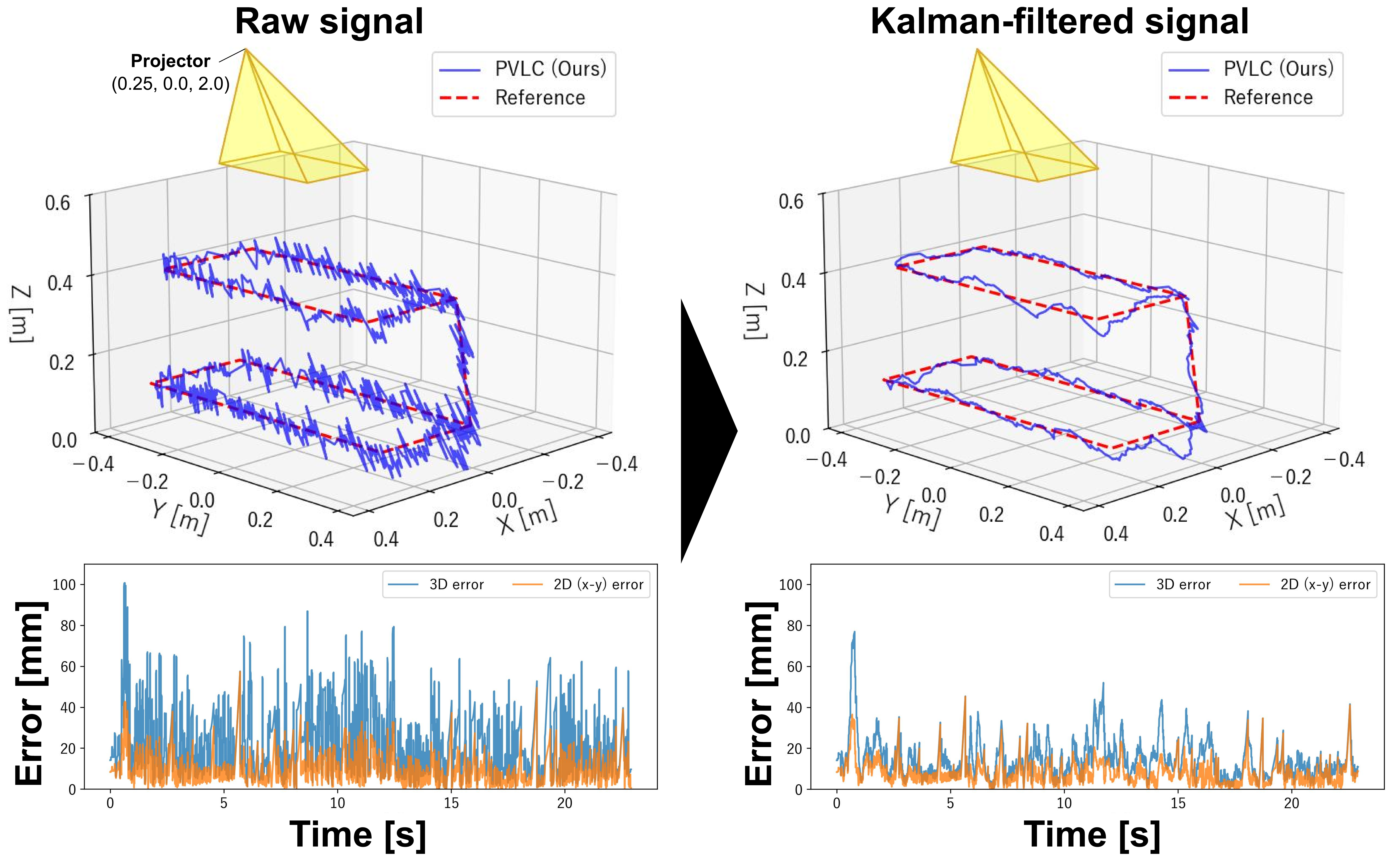}
{\textbf{Pose estimation results of the raw PVLC signal and the Kalman-filtered signal. Top row: The movement trajectory and the pose estimation results of the device.
Bottom row: The time variation of error magnitudes in 3D and 2D (x-y plane). The example was obtained with the 80 mm photosensor layout at a movement speed of 200 mm/s.}\label{fig:error_kalman}}

\Figure[t!](topskip=0pt, botskip=0pt, midskip=0pt)[width=\textwidth]{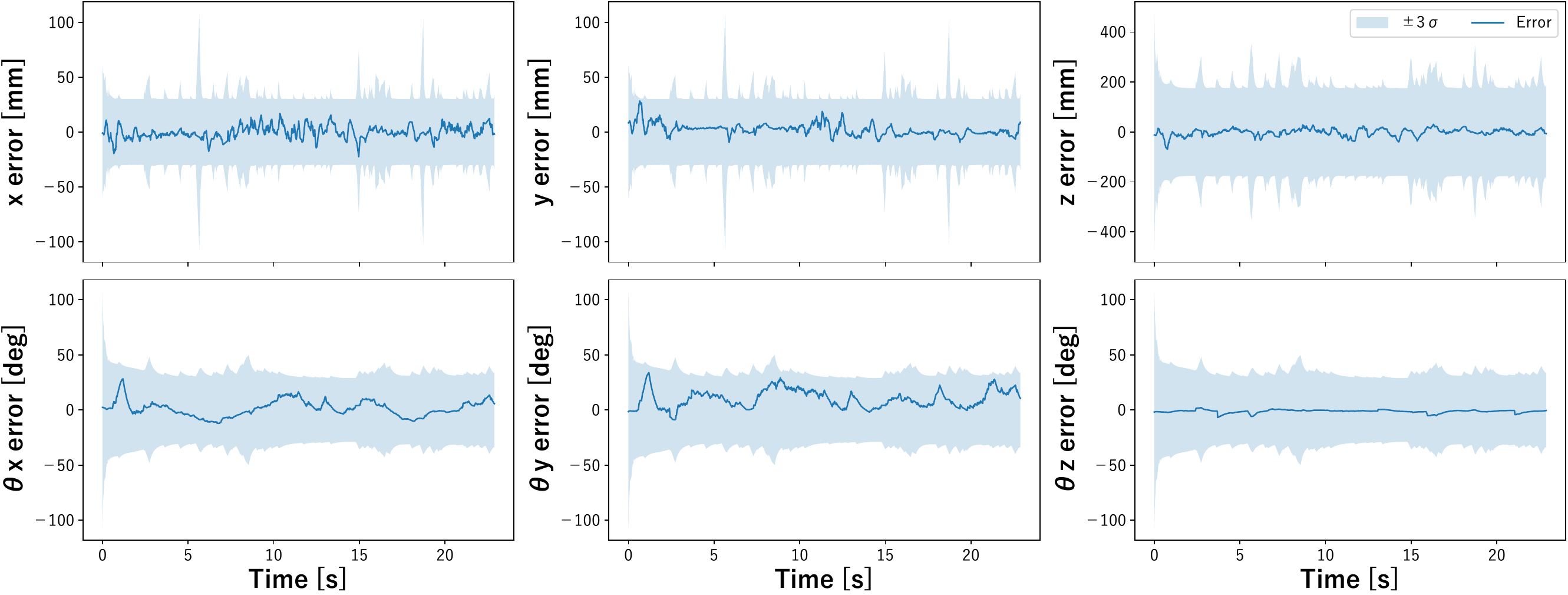}
{\textbf{Estimation error after Kalman filtering and the $\pm3\sigma$ bounds obtained from the error covariance for the position ($x$, $y$, $z$) and rotation vector ($\theta_x$, $\theta_y$, $\theta_z$) components. The example was obtained with the 80 mm photosensor layout at a movement speed of 200 mm/s.}\label{fig:kalman_3sigma}}

\section{Discussion}
\label{sec:discussion}

\subsection{Theoretical Accuracy Bounds and Error Analysis}
\label{subsec:theoretical_bounds}
To quantitatively evaluate the accuracy potential of the proposed system, we derive theoretical lower bounds on the pose estimation error. At a distance $d$ from the projector with focal length $f$, the physical size of one pixel is $d/f$. From the error propagation of the pinhole camera model~\cite{Hartley2003}, the theoretical lower bound on the 2D RMSE for $N$ photosensors is
\begin{equation}
  \sigma_{\mathrm{2D}} = \frac{d}{f} \cdot \frac{\sigma_p}{\sqrt{N}},
  \label{eq:sigma_2d}
\end{equation}
where $\sigma_p$ denotes the pixel-level decoding uncertainty, assumed to be one pixel. Similarly, from the relationship between depth and the apparent size of the photosensor layout~\cite{Hartley2003}, the depth error is
\begin{equation}
  \sigma_z = \frac{d^2}{f \cdot b} \cdot \frac{\sigma_p}{\sqrt{N}},
  \label{eq:sigma_z}
\end{equation}
where $b$ is the baseline distance between photosensors ($b = 80$ or $160$~mm in this study). The angular error per rotation axis is
\begin{equation}
  \sigma_\theta = \frac{\sigma_p \cdot d}{f \cdot b \cdot \sqrt{N}}.
  \label{eq:sigma_theta}
\end{equation}

Substituting the focal length of the projector used in this study ($f \approx 1570$), a representative distance of $d = 1.7$~m, and $N = 4$ into these expressions gives $\sigma_{\mathrm{2D}} \approx 0.54$~mm for both layouts. For $b = 80$~mm, $\sigma_{\mathrm{3D}} = \sqrt{\sigma_{\mathrm{2D}}^2 + \sigma_z^2} \approx 11.5$~mm and $\sigma_{\mathrm{angle}} = \sqrt{3}\,\sigma_\theta \approx 0.012$~rad ($0.67^{\circ}$), and for $b = 160$~mm, $\sigma_{\mathrm{3D}} \approx 5.8$~mm and $\sigma_{\mathrm{angle}} \approx 0.006$~rad ($0.34^{\circ}$).

Table~\ref{tab:bounds} compares these bounds with the experimental RMSE at the lowest speed (50 mm/s) in Section~\ref{sec:localize_dynamic}. All measured errors exceed their bounds, and the main source of the gap differs among the components.

The 3D error is dominated by its depth component. Enlarging the layout reduced the 3D error by a factor of approximately 0.58, close to the factor of 0.50 predicted by the inverse proportionality of $\sigma_z$ to $b$. If the gap came from systematic errors independent of the spacing, it would not shrink when the layout is enlarged. The measured error instead shrank at nearly the same rate as the bound. Because $\sigma_z$ is proportional to $\sigma_p / b$, this behavior indicates an effective decoding uncertainty $\sigma_p$ larger than the assumed one pixel. For both layouts, the measured error is approximately twice the bound, which corresponds to an effective uncertainty of about two pixels. Such an uncertainty can occur when the light-receiving area of a photosensor spans more than one pixel or when the pixel boundaries are blurred.

The 2D errors are larger than their bound. This bound is small because a pixel error shifts the lateral position by only $d/f$ per pixel, without the additional amplification by $d/b$ that acts on the depth. The measured 2D error is therefore sensitive to factors outside the model, such as calibration errors and the slightly oblique projection in our setup. The latter mixes part of the depth error into the $x$--$y$ plane.

The angular errors also exceed their bounds. Eq.~\ref{eq:sigma_theta} assumes that every rotation axis is constrained as tightly as the in-plane rotation. In practice, the tilt about the axes parallel to the sensor board is observable only through perspective foreshortening, which changes little near a fronto-parallel pose. The tilt estimate is therefore more sensitive to pixel errors than the bound suggests, and it can flip between the two solutions of planar pose estimation~\cite{collins2014infinitesimal}. The large SD of the angle error relative to its RMSE in Table~\ref{tab:bounds} is consistent with such intermittent flips.

\begin{table}[t]
\caption{\textbf{Theoretical lower bounds and measured pose estimation errors of the proposed PVLC at 50 mm/s. Measured values are presented as RMSE $\pm$ SD.}}
\label{tab:bounds}
\centering
\setlength{\tabcolsep}{4pt}
\begin{tabular}{cc|cc|}
     & \textbf{Layout} & \textbf{Bound} & \textbf{Measured} \\ \hline
    \textbf{3D}    & 80 mm  & 11.5 mm & $23.4 \pm 13.9$ mm \\
                   & 160 mm & 5.8 mm  & $13.5 \pm 7.3$ mm  \\ \hline
    \textbf{2D}    & 80 mm  & 0.54 mm & $8.5 \pm 5.0$ mm   \\
                   & 160 mm & 0.54 mm & $6.8 \pm 3.9$ mm   \\ \hline
    \textbf{Angle} & 80 mm  & $0.67^{\circ}$ & $21.6^{\circ} \pm 15.5^{\circ}$ \\
                   & 160 mm & $0.34^{\circ}$ & $7.4^{\circ} \pm 5.5^{\circ}$  \\
\end{tabular}
\end{table}

\subsection{Comparison with Camera-Based AR Markers}
\label{subsec:speed_dependence}
At low speeds, the proposed method was less accurate than the AR marker, particularly in orientation (Table~\ref{tab:error_dynamic}). Two factors likely contribute to this difference. First, the camera image has more pixels than the projector image. Second, the camera locates the marker corners with sub-pixel precision, whereas each photosensor receives only the integer coordinate of the pixel illuminating it. Enlarging the layout, however, reduced the error of the proposed method more than that of the AR marker. A wider spacing makes the layout span more projector pixels, which reduces the relative effect of the integer quantization (Section~\ref{subsec:theoretical_bounds}). The photosensor spacing is thus an effective design parameter for narrowing this gap.

As the speed increased, the proposed method closed this gap. Its error grew only moderately with speed, whereas that of the AR marker grew sharply. This difference can be attributed to the measurement principles of the two methods. For the AR marker, the marker moves during the exposure time of each frame, which causes motion blur and degrades corner detection. A shorter exposure time would reduce the blur, but it also reduces the light collected in each frame and increases image noise. Compensating for this loss with brighter illumination would in turn lower the contrast of the projected content. In the proposed method, each photosensor directly receives the projected light and decodes the position code of its pixel. The method therefore involves no image exposure and is inherently free from this trade-off. 

\subsection{PnP Solver Characteristics Required for This System}
Because each additional photosensor increases cost, the system should use as few photosensors as possible. From this viewpoint, P3P, which estimates pose from three points, would be advantageous. However, the P3P results in Section~\ref{subsec:exp_pnp} contained prominent outliers, which degraded its accuracy. Because P3P does not use the coplanarity of the photosensors, its solutions are more ambiguous.
On the other hand, the Homography Decomposition method and IPPE explicitly use the coplanarity constraint, and showed the best pose estimation accuracy among the compared solvers in Section~\ref{subsec:exp_pnp}.
From these points, the required PnP solver for this system should be capable of ensuring accuracy by exploiting geometric constraints even with a minimal number of correspondence points. Additionally, since all solvers had sufficiently fast processing times of less than 1 ms, an increase in computational load to maintain accuracy is within an acceptable range.

\subsection{Robustness to Light Disturbance}
When this method is used in actual operation under various environments, it is necessary to consider degradation and false detection of received signals due to disturbance factors such as ambient light and light occlusion.

Regarding ambient light, the current implementation already incorporates a high-pass filter in the signal processing circuit (see Section~\ref{sec:implementation}), which suppresses low-frequency ambient light components from typical indoor lighting such as fluorescent lamps. All experiments in this study were conducted with the room lights on (Section~\ref{sec:experiments}), and as their results show, the system operates stably under normal indoor lighting. Furthermore, by employing wavelengths underrepresented in ambient illumination, such as infrared, for the projector light signal, or by installing optical bandpass filters on the photosensor side, the system can be extended to handle a wider range of lighting conditions.
For light occlusion, a redundant photosensor configuration allows the remaining photosensors to supply the signals when some of them are blocked. Other sensors such as inertial measurement units (IMUs) can also temporarily supplement position estimation.

\section{Limitations and Future Work}
\label{sec:limitations}
This study primarily focuses on demonstrating the feasibility of 3D pose estimation using PVLC. This section describes the current limitations of the proposed method and the corresponding directions for future work.

\subsection{Movement Speed}
\label{subsec:lim_speed}
The pose estimation experiments during device movement (Section~\ref{sec:localize_dynamic}) were conducted at speeds up to 400 mm/s. More agile drones or robots move even faster. At high speeds, a photosensor crosses multiple pixels during one cycle of the position code. We expect this to increase decoding errors. Suppressing this effect by shortening the code cycle or by motion-aware decoding is a direction for future work.

\subsection{Device Size and Geometry}
\label{subsec:lim_geometry}
In this study, we evaluated photosensor layouts of 80 mm and 160 mm squares. Applying the proposed method to small devices requires a more compact layout. However, the theoretical bounds in Section~\ref{subsec:theoretical_bounds} (Eqs.~\ref{eq:sigma_z} and~\ref{eq:sigma_theta}) show that the depth and angular errors increase in inverse proportion to the baseline $b$. The measured errors for the two layouts followed this trend. A compact device therefore needs measures that offset the accuracy loss caused by miniaturization, such as a higher-resolution projector or additional photosensors. Fusing the pose estimates with inertial measurements through an Extended Kalman Filter (EKF) would also help. This fusion would be most effective along the $z$-axis, where the error is largest and increases as the layout shrinks.

The Homography Decomposition method and IPPE, which assume coplanar photosensors, achieved the highest accuracy in Section~\ref{subsec:exp_pnp} with only four photosensors. However, non-planar devices, such as those with curved surfaces, must instead use solvers that do not assume coplanarity, such as P3P and EPnP. In our experiments, these solvers were less accurate than the planar solvers. Non-planar devices would therefore likely need more photosensors to reach comparable accuracy. Fusion with other sensors such as IMUs is another option for such geometries.

\subsection{Operating Area}
\label{subsec:lim_area}
The operating area of the proposed method is inherently limited to the projection volume of the projector. This study, however, targets devices that operate in coordination with projected images and therefore does not consider operation outside the projection volume. To extend the operating area, one can steer the projection direction~\cite{Hiraki2019-fn} or use multiple projectors.

\section{Conclusion}
\label{sec:conclusion}
In this paper, we proposed a 3D pose estimation method using PVLC based on the PnP problem solution to obtain the 3D position of devices coordinated with projected images.
We evaluated the pose estimation accuracy of the proposed method under different movement speeds and photosensor layout sizes, in comparison with a camera-based AR marker method.
As a result, we confirmed that the proposed method achieves cm-order pose estimation accuracy with a latency of 8 ms, and that its error is less sensitive to movement speed than that of the AR marker. These properties enable effective coordination between projected images and device motion, facilitating the construction of highly immersive augmented reality systems.

As future work, we plan to investigate higher-resolution projectors to further improve pose estimation accuracy and robustness. Additionally, fusing the proposed pose estimation with inertial measurements via an Extended Kalman Filter is a promising direction for enhancing estimation stability under dynamic motion.
Finally, enabling the PnP solver to run on compact and lightweight computing environments will improve device portability and broaden the range of practical applications.

\bibliographystyle{IEEEtran}
\bibliography{access2025_3dpvlc}

\begin{IEEEbiography}[{\includegraphics[width=1in,height=1.25in,clip,keepaspectratio]{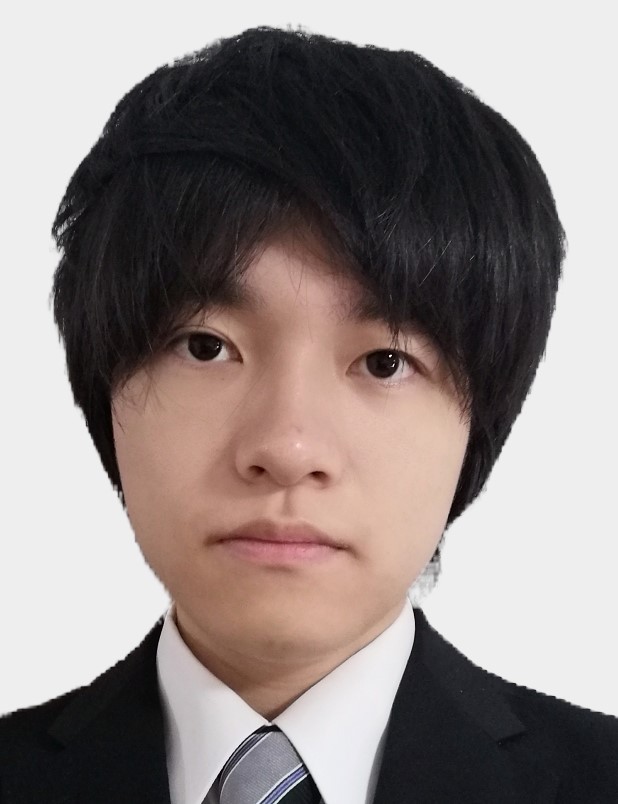}}]{Michikuni Eguchi}
received the B.E. and M.I. degrees from the University of Tsukuba, Japan in 2023 and 2025, respectively. Currently, he is a Ph.D student at the University of Tsukuba, Japan. His research interests include robotics, motion planning, and haptics.
\end{IEEEbiography}

\begin{IEEEbiography}[{\includegraphics[width=1in,height=1.25in,clip,keepaspectratio]{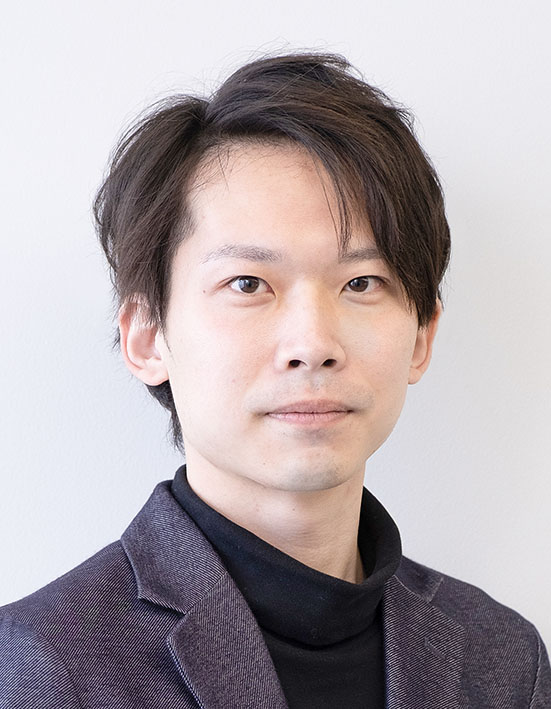}}]{Takefumi Hiraki}
received his B.S., M.S., and Ph.D. degrees from the University of Tokyo, Japan in 2014, 2016, and 2019, respectively. From 2019, he worked as a Japan Society for the Promotion of Science (JSPS) Research Fellow at the Graduate School of Engineering Science at Osaka University. From 2021, he served as an assistant professor at the Institute of Library, Information and Media Science, University of Tsukuba. Currently, he is an Associate Professor at the Institute of Library, Information and Media Science, University of Tsukuba, and Director of Cluster Metaverse Lab, Cluster, Inc. His research interests include augmented reality, metaverse, human-computer interaction, haptic interfaces, and soft robotics.
\end{IEEEbiography}

\EOD

\end{document}